\documentclass{article}

\usepackage{microtype}
\usepackage{graphicx}
\usepackage{subcaption}
\usepackage{booktabs} 
\usepackage{multirow}

\usepackage{hyperref}

\usepackage[preprint]{icml2026}

\usepackage{amsmath}
\usepackage{amssymb}
\usepackage{mathtools}
\usepackage{amsthm}

\usepackage[capitalize,noabbrev]{cleveref}

\theoremstyle{plain}

\theoremstyle{definition}

\theoremstyle{remark}

\usepackage[textsize=tiny]{todonotes}

\icmltitlerunning{Vision-Language Models Unlock Task-Centric Latent Actions}

\begin{document}

\twocolumn[
  \icmltitle{Vision-Language Models Unlock Task-Centric Latent Actions}



  \icmlsetsymbol{equal}{*}
  \icmlsetsymbol{airiwork}{†}
  
  \begin{icmlauthorlist}
    \icmlauthor{Alexander Nikulin}{airi,msu}
    \icmlauthor{Ilya Zisman}{hum,airiwork}
    \icmlauthor{Albina Klepach}{airi}
    \icmlauthor{Denis Tarasov}{airi,airiwork}
    \icmlauthor{Alexander Derevyagin}{airi}
    \icmlauthor{Andrei Polubarov}{airi,skol,ras}
    \icmlauthor{Lyubaykin Nikita}{airi,inno}
    \icmlauthor{Vladislav Kurenkov}{airi,inno}
  \end{icmlauthorlist}

    \icmlaffiliation{airi}{AIRI}
    \icmlaffiliation{hum}{Humanoid}
    \icmlaffiliation{msu}{MSU}
    \icmlaffiliation{inno}{Innopolis University}
    \icmlaffiliation{skol}{Skoltech}
    \icmlaffiliation{ras}{Research Center for Trusted Artificial Intelligence, ISP RAS}
  
  \icmlcorrespondingauthor{Alexander Nikulin}{nikulin@airi.net}
  \icmlkeywords{Machine Learning, ICML, LAM, RL, Robotics}

  \vskip 0.3in
]



\printAffiliationsAndNotice{\icmlWorkDoneAtAIRI}

\begin{abstract}
    Latent Action Models (LAMs) have rapidly gained traction as an important component in the pre-training pipelines of leading Vision-Language-Action models. However, they fail when observations contain action-correlated distractors, often encoding noise instead of meaningful latent actions. Humans, on the other hand, can effortlessly distinguish task-relevant motions from irrelevant details in any video given only a brief task description. In this work, we propose to utilize the common-sense reasoning abilities of Vision-Language Models (VLMs) to provide promptable representations, effectively separating controllable changes from the noise in unsupervised way. We use these representations as targets during LAM training and benchmark a wide variety of popular VLMs, revealing substantial variation in the quality of promptable representations as well as their robustness to different prompts and hyperparameters. Interestingly, we find that more recent VLMs may perform worse than older ones. Finally, we show that simply asking VLMs to ignore distractors can substantially improve latent action quality, yielding up to a six-fold increase in downstream success rates on Distracting MetaWorld.
\end{abstract}

\section{Introduction}

\begin{figure}[ht]
  \vskip 0.2in
  \begin{center}
    \centerline{\includegraphics[width=0.9\columnwidth]{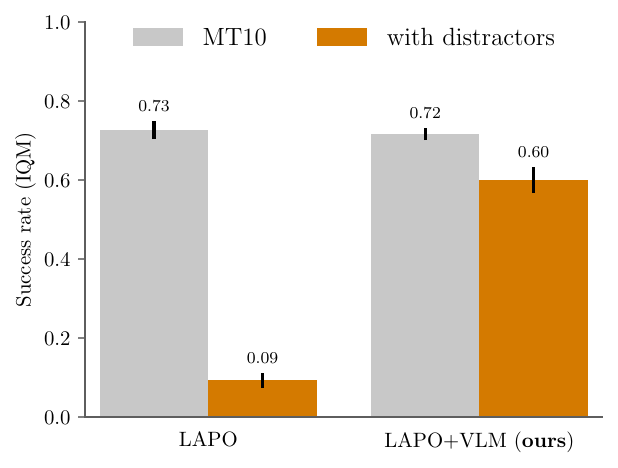}}
    \caption{
      \textbf{Main result}. Success rate on MetaWorld-10 benchmark for LAPO and proposed LAPO+VLM (Molmo), which uses promptable representations. We use three random seeds and report IQM and $95\%$-CI based on stratified bootstrapping, following the \citet{agarwal2021deep}. See \cref{sec:sc-rate} for full results.
    }
    \label{fig:main-result}
  \end{center}
  \vskip -0.4in
\end{figure}


Latent action models (LAMs) \citep{schmidt2023learning, ye2024latent} have quickly become integral to the pre-training pipelines of leading Vision–Language–Action (VLA) models \citep{bjorck2025gr00t, bu2025agibot, zhong2025survey, bruce2024genie, jang2025dreamgen}. By inferring compact, semantically meaningful latent action representations at scale, LAMs mitigate the scarcity of high-quality action-labeled data, removing the data bottleneck in embodied AI and robotics \citep{mccarthy2025towards}. Unfortunately, early LAMs that gained recognition \citep{schmidt2023learning, ye2024latent, chen2024igor, gao2025adaworld} were trained on relatively clean datasets, where changes between observations could be explained almost entirely by ground-truth actions, such as in static scenes with a single manipulator. In contrast, real-world data often contains numerous action-correlated distractors, including background human movement or other spurious correlations. 

However, as shown by \citet{nikulin2025latent, zhang2025latent, klepach2025object}, without explicit supervision, LAMs struggle to disentangle controllable changes from noise, completely failing to produce meaningful latent actions in the presence of action-correlated distractors. Humans, however, interpret the world through semantics rather than raw pixels, and with only a brief task description can easily separate task-relevant features from irrelevant details. Wouldn't it also be convenient to simply ask LAM to focus on the relevant features, e.g. robotic arm, and ignore any other details?


\begin{figure*}[t]
\vskip 0.2in
    \centering
    \includegraphics[width=0.95\textwidth]{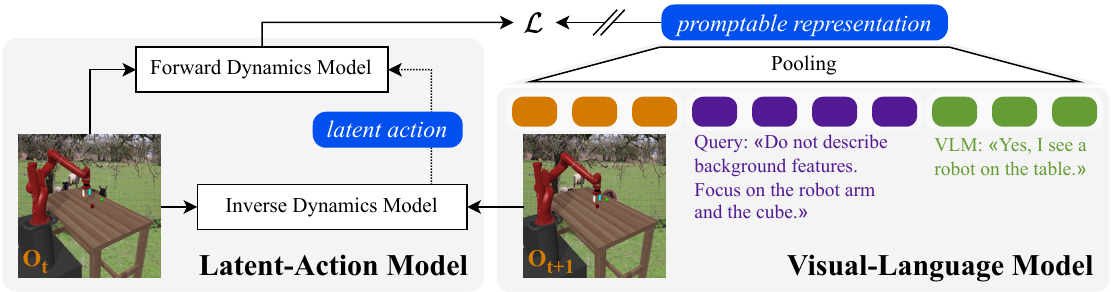}
    \caption{Visualization of the task-relevant promptable representations extraction from the VLMs and their subsequent use as targets during latent action learning.}
    \label{fig:scheme}
\vskip -0.2in
\end{figure*}

Inspired by the work of \citet{chen2024vision} on promptable representations, we propose to utilize the Vision-Language Models (VLMs) common-sense reasoning abilities as an unsupervised approach for effectively separating controllable changes from noise, thereby restoring the LAM's ability to recover ground-truth actions even in the presence of distractors. Using a controlled Distracting MetaWorld \citep{yu2020meta} setup, we show that representations produced by prompting VLMs to focus on task-specific details can serve as an effective target for LAMs in the presence of distractors, increasing the success rate six-fold (see \Cref{fig:main-result}). 

As a motivation for our approach, we begin from a simple demonstration experiment, showing that limitations of LAM can be mitigated with the clean target (\Cref{sec:lapo-twin}). We then conduct large-scale benchmarking of different VLMs, comprising over $29$k+ experiments, to assess their effectiveness at providing promptable representations (\Cref{sec:vlm-bench}), revealing substantial variation in quality and robustness to hyperparameters. In addition, we show that VLMs significantly outperform self-supervised methods, such as DINOv2 \citep{oquab2023dinov2} or CLIP \citep{radford2021learning}, confirming our hypothesis that language based filtering is essential for performance. Finally, using the best setup found, we demonstrate that, without any supervision, promptable representations can significantly improve latent action quality and downstream performance (\Cref{sec:sc-rate}), outperforming OTTER \citep{huang2025otter} and UniVLA \citep{bu2025univla} baselines.

\section{Preliminaries}

\textbf{Problem setting.} We consider a setting of offline imitation learning from observation \citep{liu2018imitation, torabi2019recent}. Our goal is to pre-train a policy $\pi(o | a)$, given a large dataset of expert trajectories $\mathcal{D} \coloneq \{ (o_i^n) \}_{i=1}^\tau$, containing observations but not actions (e.g. videos), and a limited number of real action labels. Ideally, the pre-trained agent should achieve maximum performance (e.g. success rate) in the environment after fine-tuning with a minimum amount of action-labeled data. The commonly considered ratio of labeled to unlabeled data is around $2-10\%$ in the existing work \citep{zheng2023semi, nikulin2025latent}, while in our experiments, we consider a ratio as low as $<1\%$.


\textbf{Promptable representations.} We follow the \citet{chen2024vision} and define promptable representations \emph{simply as a process of obtaining observation embeddings from the VLMs given a task-specific prompt and some extraction and aggregation strategy} (see \Cref{fig:scheme}). We obtain such representations from the last and next-to-last layers \citep{chen2024vision}. In contrast to the \citet{chen2024vision, huang2025otter} we cannot learn pooling from the data to better predict true actions or obtain better reward. Thus, we experiment only with simple fixed strategies, such as taking the mean over all output embeddings or taking only the embedding of the last token from either prompt or the generated answer. Thus, the final promptable representations of an observation $o_t$ is always just a single vector $s_t \in R^D$, where $D$ is a VLMs hidden size.

\textbf{Latent action models.}  Given the dataset of observations $\mathcal{D} \coloneq \{ (o_i^n) \}_{i=1}^\tau$, LAMs \citep{rybkin2018learning, edwards2019imitating,schmidt2023learning} try to infer latent actions $z_t$ such that they are maximally predictive of observed transitions $(o_t, o_{t+1})$ while being minimal \citep{schmidt2023learning}, i.e. describe changes only relevant to control. After pre-training, LAM is used to supply 
latent actions for behavioral cloning (BC) on unlabeled dataset to obtain useful behavioral priors. As a final stage, decoder is trained to map from latent to ground-truth actions on a small labeled dataset.

Modern LAMs \citep{bruce2024genie, ye2024latent, cui2024dynamo, chen2024moto, chen2024igor, gao2025adaworld} mostly follow the same high-level architecture introduced by LAPO \citep{schmidt2023learning}, which uses a combination of inverse (IDM) and forward (FDM) dynamics models to infer latent actions. Given a transition $(o_t, o_{t+1})$, IDM first infers latent action $z_t \sim f_{\text{IDM}}(\cdot | o_t, o_{t+1})$, which FDM further uses to predict the next observation $\hat{o}_{t+1} \sim f_{\text{FDM}}(\cdot | o_t, z_t)$. Both models are trained jointly to minimize the loss
$
\mathcal{L}_\mathrm{MSE} = \mathbb{E}_{(o_t, o_{t+1}) \sim \mathcal{D}} \left[ \left\| f_\mathrm{FDM}(f_\mathrm{IDM}(\boldsymbol{o}_t, \boldsymbol{o}_{t+1}), \boldsymbol{o}_t) - \boldsymbol{o}_{t+1} \right\|^2 \right]
$.

To avoid shortcut learning, i.e. IDM just passing $o_{t+1}$ as is, the information bottleneck is usually applied to latent actions, with quantization via VQ-VAE \citep{van2017neural} as a popular choice \citep{schmidt2023learning, bruce2024genie, chen2024moto}. There is, however, growing evidence that quantization is harmful \citep{nikulin2025latent, liang2025clam, yang2025learning} and simply restricting dimensionality is enough.


\begin{figure}[t]
\vskip 0.2in
    \centering
    \includegraphics[width=0.8\columnwidth]{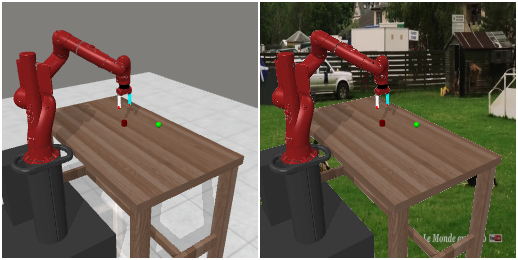}
    \caption{Visualization of observations with and without distractors in our modification of MetaWorld environment.}
    \label{fig:metaworld}
\vskip -0.2in
\end{figure}

\begin{figure*}[t]
    \begin{subfigure}[b]{0.33\textwidth}
        \centering
        \centerline{\includegraphics[width=\columnwidth]{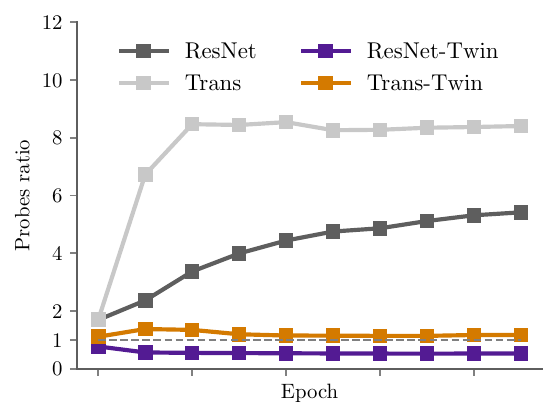}}
        \caption{w/ to wo/ distractors probe ratio}
        \label{fig:baselines-probe-ratio}
    \end{subfigure}
    \begin{subfigure}[b]{0.33\textwidth}
        \centering
        \centerline{\includegraphics[width=\columnwidth]{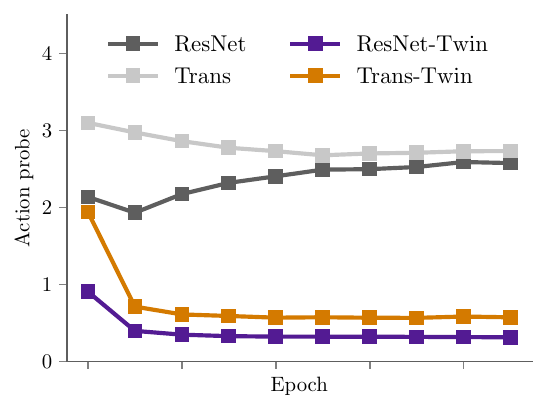}}
        \caption{Probes w/ distractors}
        \label{fig:baseliens-probe-dist}
    \end{subfigure}
    \begin{subfigure}[b]{0.33\textwidth}
        \centering
        \centerline{\includegraphics[width=\columnwidth]{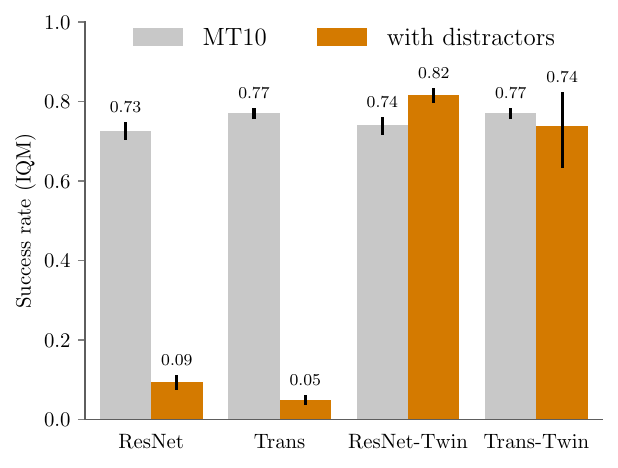}}
        \caption{Success rate on MT10}
        \label{fig:baselines-succces-rate}
    \end{subfigure}
    \caption{Demonstration that quality of latent actions learned by LAPO completely degrades in the presence of distractors, which results in almost zero success rate. We show that with the ideal target for FDM, which perfectly disentangles controllable features from the noise, performance may be restored, serving as a main motivation for us to explore promptable representations. Action probe represents MSE of a linear probe trained to predict real actions from latent actions. See \Cref{sec:setup} for detailed explanation. We use three random seeds and report IQM and $95\%$-CI based on stratified bootstrapping, following the \citet{agarwal2021deep}. See \Cref{sec:lapo-twin} for details.}
    \label{fig:baselines}
\end{figure*}

\begin{figure}[t]
    \centering
    \includegraphics[width=\columnwidth]{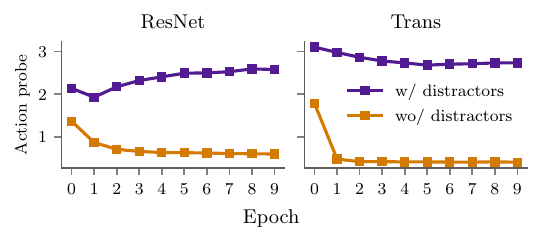}
    \caption{Baseline LAPO action probes on MT10. Averaged over 3 random seeds. Action probe represents MSE of a linear probe trained to predict real actions from latent actions. See \Cref{sec:setup} for detailed explanation.}
    \label{fig:lapo-resnet-trans-lim}
\vskip -0.2in
\end{figure}

\section{Related Work}
\label{sec:related}

\textbf{Latent action learning.} \citet{edwards2019imitating} was among the first to introduce the problem of recovering latent actions, demonstrating limited success on simple tasks. LAPO \citep{schmidt2023learning} substantially simplified the approach, eliminated key scalability barriers, and achieved high success in the challenging, procedurally generated tasks, helping to popularize LAMs. Subsequent work further scaled latent action learning to larger models and datasets, and to increasingly complex and diverse robotics domains \citep{bruce2024genie, cui2024dynamo, ye2024latent, chen2024moto, chen2024igor, jang2025dreamgen, bjorck2025gr00t}.

Recent studies \citep{nikulin2025latent, klepach2025object, zhang2025latent} have highlighted the failure of LAMs in the presence of action-correlated distractors. Both \citet{nikulin2025latent} and \citet{zhang2025latent} proposed incorporating supervision from a small number of true actions during LAM training to help identify controllable changes. While this approach is effective, it is not generalizable, since in many domains, such as egocentric human videos \citep{yang2025egovla}, true actions cannot be obtained in a practical manner. In contrast, we propose leveraging promptable representations from VLMs as an unsupervised approach to separate controllable changes from noise. We empirically demonstrate that, without any supervision, this enables LAMs to recover the performance in the presence of distractors (see \Cref{fig:main-result}).

The work most closely related to ours is UniVLA \citep{bu2025univla}, which also focuses on task-centric latent actions. UniVLA uses DINOv2 features, which are intended to provide priors that better capture task-relevant information. In addition, it proposes a two-stage LAM training pipeline that incorporates language task instruction embeddings as inputs to the LAM in the first stage. However, UniVLA design has important limitations. While conditioning the LAM on task instructions may help separate task-relevant videos from noise at a high level, it does not provide the per-step learning signal required to accurately recover low-level ground-truth actions. As a result, in the single-task learning setting the task instruction remains a constant vector, and UniVLA effectively reduces to LAPO, leading to noisy latent actions. In contrast, we provide the task instruction to the VLM, and while the instruction itself is constant, the VLM produces different language-conditioned representations for each observation, which LAM can use as prediction targets (see \Cref{fig:scheme}). Thus, our approach is simpler, does not require multiple stages, provides clean targets at each step, and remains effective in the single-task regime, significantly outperforming UniVLA (see \Cref{sec:sc-rate}) and DINOv2 (see \Cref{sec:vlm-bench}).

\textbf{Promptable representations.} Recently, \citet{chen2024vision} proposed using VLM embeddings generated with task-specific prompts to extract improved state representations, which enhance sample efficiency in online reinforcement learning. Our work is heavily inspired by this approach, and we closely follow their methodology for extracting promptable representations. Our contribution, however, is a new setting and novel empirical insights. Results of \citet{chen2024vision} do not guarantee the minimality of representations and, as a consequence, success in latent action learning. In contrast, our work shows that VLMs differ substantially in their zero-shot suitability for LAMs, and that selecting an appropriate VLM can make LAMs effective in the presence of distractors, increasing success rates by up to sixfold, a result that has not been documented in prior work.



\section{Experimental Setup}

\label{sec:setup}
\textbf{Environments and datasets.} We use MetaWorld \citep{yu2020meta} Multi-Task 10 (MT10) as our primary benchmark, as it allows us to conduct experiments in controlled setting while being lightweight enough for exhaustive experimentation with VLMs under limited resources. We modify MetaWorld to include distracting videos in the background, using the DAVIS \citep{pont20172017} videos, which depict different real world scenes, including humans. We also move the default camera position farther back and remove borders around the table to include more of the background video in the observation, making latent action learning more challenging. See \Cref{fig:metaworld} for a visualization. 

We follow the standard three-stage pipeline \citep{schmidt2023learning, ye2024latent, nikulin2025latent}: (1) pre-train the LAM, (2) train behavioral cloning (BC) agent on latent actions, and (3) train a decoder head on a small number of true-action labels. For each task, we collect 5k trajectories from the scripted experts provided by MetaWorld and up to 16 additional labeled trajectories for the final stage, which is less than $1\%$ of the full datasets.

\textbf{Latent action model architecture.} We use the architecture proposed by \citet{schmidt2023learning}, omitting action quantization \citep{nikulin2025latent, liang2025clam, yang2025learning}. We use frame stacking, but only in IDM, while FDM uses only the current frame and latent action to predict the next frame, as in \citet{chen2024igor}. Other than that, in our main experiments, we do not use any improvements upon LAPO (unless explicitly stated otherwise), such as augmentations or multi-step predictions in FDM \citep{nikulin2025latent, chen2024igor, ye2024latent, cui2024dynamo}, to eliminate possible confounders on latent action quality. We use ResNet \citep{he2016deep} or ViT \citep{dosovitskiy2020image} as IDM and FDM backbones. When predicting in the latent space instead of images, we follow \citet{nikulin2025latent} and use multiple MLP blocks similar to those used in Transformers \citep{vaswani2017attention}. For action decoder head, we use a small three-layer MLP. For actor itself, we always used ResNet \citep{he2016deep}. For hyperparameters see \Cref{app:hp} in the Appendix.

\textbf{Evaluation.} For evaluation we use \textbf{action probe} and \textbf{success rate}. Specifically, we train linear probes to predict real actions from the latent ones during LAM training, while stopping the gradient through the latent actions. The final MSE serves as our quality metric, as it indicates whether the latent actions encode the real ones. This metric is also used for hyperparameter tuning, which may be impractical in real-world settings but allows us to estimate the upper-bound performance of each method for fair comparison. 

However, as \citet{nikulin2025latent} notes, linear probing has a key limitation: it can reveal whether true actions are present in the latent space, but it does not guarantee minimality, meaning that exogenous noise may still be encoded. Thus, to measure the true usefulness of pretraining with latent actions, we evaluate the success rate in the environments after fine-tuning on true action labels.


\section{The Importance of Right Target}
\label{sec:lapo-twin}

\begin{figure*}[t]
    \vskip 0.2in
    \begin{subfigure}[b]{0.49\textwidth}
        \centering
        \centerline{\includegraphics[width=\textwidth]{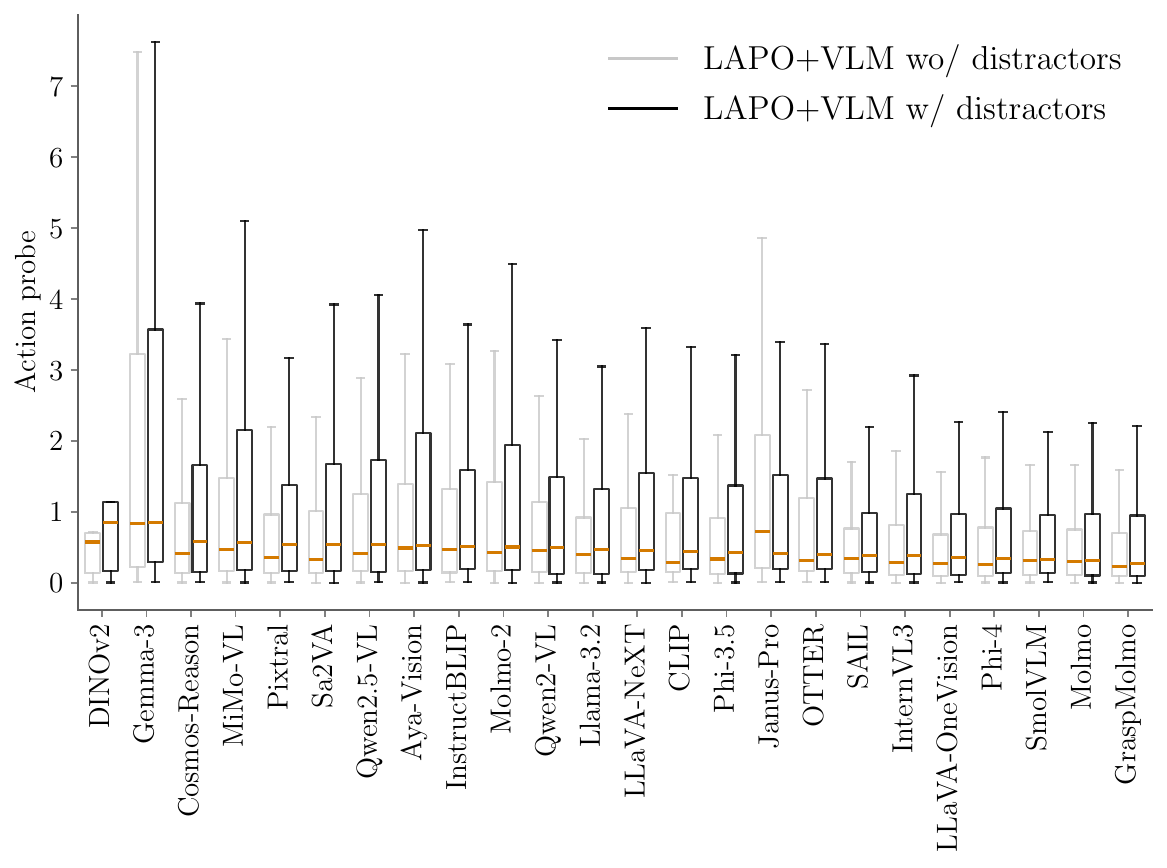}}
        \caption{Aggregation over all hyperparameters explored}
        \label{fig:vlm-benchmark-all}
    \end{subfigure}
    \hfill
    \begin{subfigure}[b]{0.49\textwidth}
        \centering
        \centerline{\includegraphics[width=\textwidth]{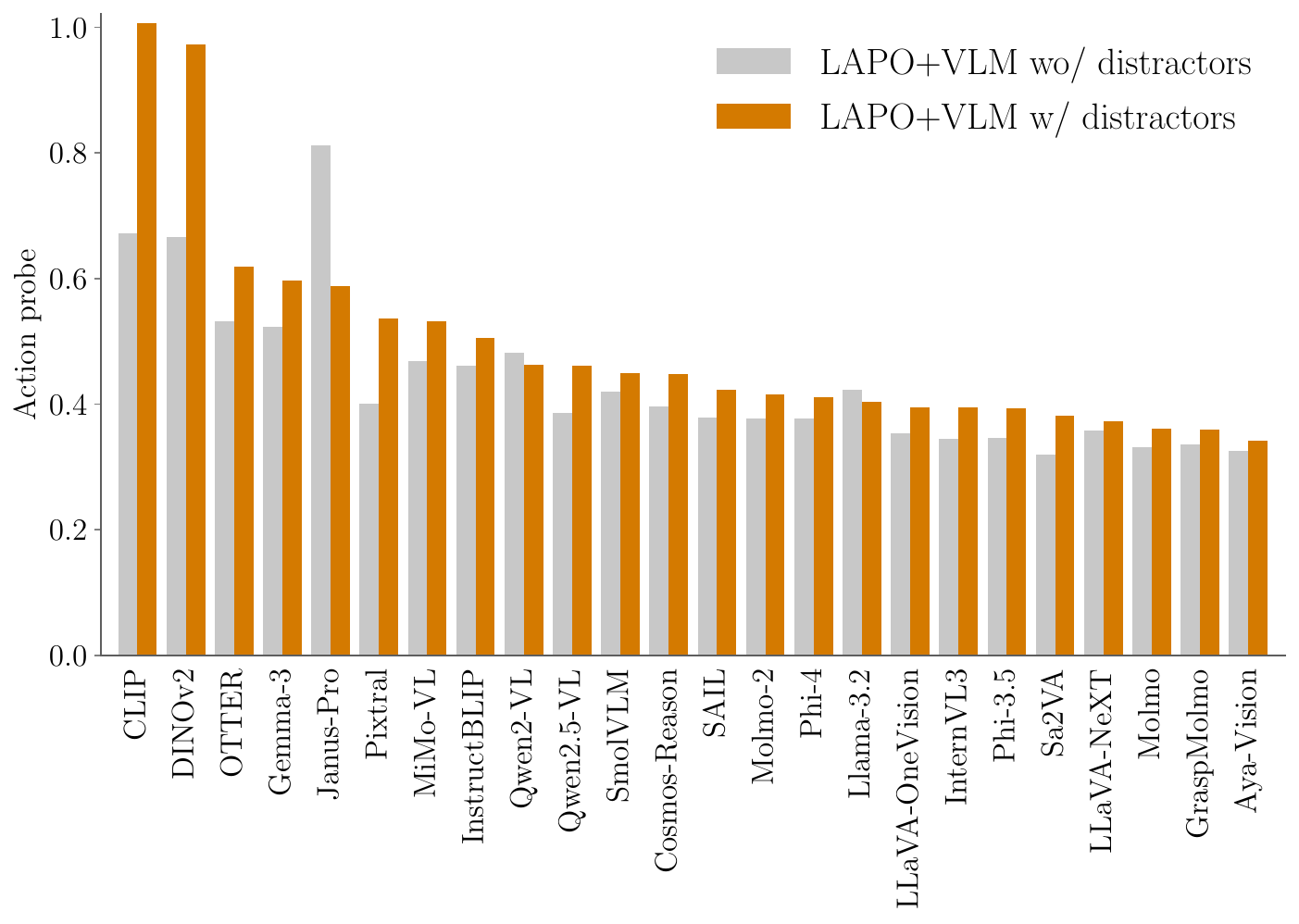}}
        \caption{Aggregation over the best hyperparameters}
        \label{fig:vlm-benchmark-best}
    \end{subfigure}
    
    \caption{Benchmarking the effectiveness of promptable representations provided by different VLMs for latent action learning on all tasks from MT10. Action probe represents MSE of a linear probe trained to predict real actions from latent actions. See \Cref{sec:setup} for detailed explanation. Overall, all VLMs provide some improvement over LAPO, with Molmo performing the best and Gemma-3 the worst. For details and exact experimental protocol see \Cref{sec:vlm-bench}. We additionally provide the ranking for each combination of hyperparameters in the \Cref{fig:vlm-hp-all} in the \Cref{app:add-figs}.}
    \label{fig:vlm-benchmark}
\end{figure*}

We begin with a simple demonstration that the limitation of LAMs in the presence of distractors arises entirely from the poor FDM target (e.g. next frame in pixel space), rather than from flaws in the overall idea or architecture. By construction, latent actions are optimized to maximally explain the dynamics. Therefore, the root of the failure to recover true actions lies in the dynamics we predict, which is directly determined by the target in FDM: $\hat{o}_{t+1} \sim f_{\text{FDM}}(\cdot | o_t, z_t)$. What would be the ideal target for FDM? And if it exists, what would be the final performance? Could LAM possibly recover the ground-truth actions despite distractors in the input observations?

\textbf{Setup.} To answer these questions we construct a special dataset with twin observations for each task: during data collection we render and save same observation with and without distractors. Next, during training we feed observations with distractors as inputs to IDM and FDM, but as the target for FDM we use next observation without distractors. As the actual controllable changes are preserved (the underlying state is the true next state), it serves as a target with ideal disentanglement of controllable features from exogenous noise (see \Cref{fig:metaworld}). To show that existing limitations are agnostic to the architecture of FDM and IDM, we explore both ResNet \citep{schmidt2023learning} and transformer \citep{bruce2024genie, ye2024latent} backbones.

\textbf{Results.} First of all, as can be seen in \Cref{fig:lapo-resnet-trans-lim}, we confirm that simply adding distractors results in complete degradation of latent actions quality regardless of backbone used. This subsequently leads to almost zero success rate after fine-tuning on true actions (see \Cref{fig:baselines-succces-rate}). Ideally, probes should be close to each other, as real underlying actions are identical between both settings. Next, in \Cref{fig:baselines-probe-ratio,fig:baseliens-probe-dist} we show the effect of using perfect targets during LAPO training (with -Twin postfix). To better illustrate the trend, in \Cref{fig:baselines-probe-ratio} we report the ratio of probes with and without distractors for each method. With the ideal target action probes immediately drop to the level of LAPO without distractors, and ratio converges to one.  Finally, improvement in latent action quality directly results in leveling success rates (see \Cref{fig:baselines-succces-rate}). Overall, this result supports that the right target is the key to unlock latent action learning in the presence of distractors. Although these experiments may seem obvious in hindsight, they allow us to convey a key empirical observation, one that provides the same intuition that originally led us to explore promptable representations. 



\section{The Promise of Promptable Representations}
\label{sec:vlm-bench}

Our main hypothesis is that representations provided by VLMs can serve as an effective, unsupervised way to disentangle controllable features from the noise. As we demonstrated in the previous section, it would be enough to unlock latent action learning in the presence of distractors. 

Modern VLMs easily identify the robotic arm location in the image (like \Cref{fig:metaworld}) and describe it in detail, even in the presence of background noise. However, the ability to generate valid text does not necessarily imply that the underlying embeddings are suitable for our purposes. For a representation to serve as an effective target for LAM, it should (1) contain task-centric visual information, (2) be minimal by filtering out visual details irrelevant to the prompt, and (3) remain consistent across observations to mimic changes caused by real actions. Unfortunately, current VLMs are known to struggle with visual focus \citep{rahmanzadehgervi2024vision, sim-etal-2025-vlms} and pixel-level understanding \citep{gou2024well, dahou2025vision, li2025lost}. Given these limitations, we begin by benchmarking a wide variety of modern VLMs to assess their viability. As a baseline, we included representations from self-supervised methods such as CLIP \citep{radford2021learning} and DINOv2 \citep{oquab2023dinov2}, which are not promptable but were pre-trained on large amounts of visual data. Based on this benchmark, we then select the most effective VLM along with the best hyperparameters (e.g., prompt, aggregation strategy, and others).

\begin{figure*}[t]
    \vskip 0.2in
    \centering
    \begin{minipage}[t]{\columnwidth}
        \centering
        \includegraphics[width=\linewidth]{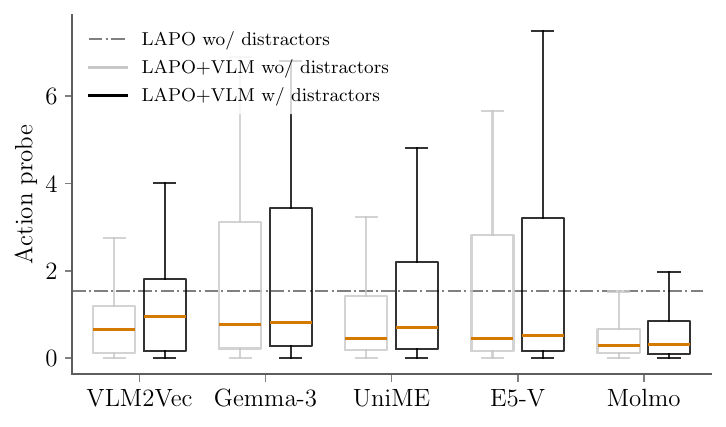}
        \caption{Benchmarking the effectiveness of promptable representations provided by recent \emph{embedding} VLMs for latent action learning on all tasks from MT10. Overall, embedding VLMs, despite their promise, do not deliver any substantial gains compared to traditional VLMs, such as Molmo. We include Gemma-3 and Molmo results here for convineince.}
        \label{fig:emb-vlm}
    \end{minipage}
    \hfill
    \begin{minipage}[t]{\columnwidth}
        \centering
        \includegraphics[width=\linewidth]{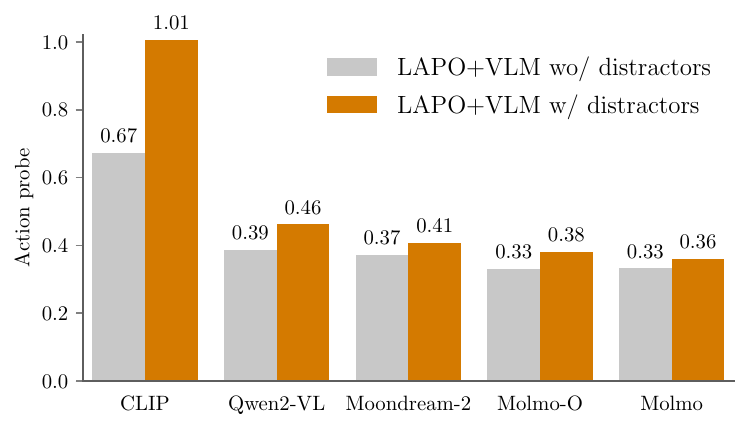}
        \caption{Molmo performance investigation, using aggregation over the best hyperparameters. We benchmark both Molmo versions, which both use CLIP but differ in LLM backbones (OLMo vs Qwen2), as well as Qwen2-VL. Since both Molmo share the pre-training data but differ in architecture, we conclude that the likely source of their superior performance lies in the data.}
        \label{fig:molmo-inv}
    \end{minipage}
\end{figure*}

\textbf{Setup.} Conducting large scale  VLMs evaluation on the full datasets would be prohibitively expensive. Instead, we train LAPO+VLM on a small subsets of 64 trajectories  instead of full 5k for each dataset and measure the resulting latent action quality. As we will show in \Cref{sec:sc-rate}, hyperparameter rankings obtained this way transfer reasonably well to the full dataset performance. For each VLM, we evaluated eight prompt variants designed in different styles to exploit diverse VLM capabilities (e.g., CLIP-style captions, pointing, segmentation; see \Cref{app:tab:prompts} in \Cref{app:vlm-list}). We further varied the source of representations (last vs. next-to-last layer, prompt vs. generated embeddings) and the aggregation strategy (averaging vs. last non-padding token). This yields 64 runs per VLM, per task, per dataset. The full list of VLMs, including exact model names, sizes, and prompt templates, is provided in \Cref{app:vlm-list}. 


\textbf{Benchmarking VLMs.} We summarize our main benchmarking results in \Cref{fig:vlm-benchmark} and provide full per-hyperparameter rankings in \Cref{fig:vlm-hp-all} (\Cref{app:add-figs}). 
As can be seen in \Cref{fig:vlm-benchmark-all}, overall all VLMs provide some degree of improvement over LAPO in terms of the median action probe. However, some of them, especially Molmo \citep{deitke2025molmo}, are generally preferable and have lower variance, indicating higher robustness to different hyperparameters. In \Cref{fig:vlm-benchmark-best} we visualize ranking by averaging best scores for each task. While this changes ranking a bit, we still observe that among VLMs Gemma-3 \citep{team2025gemma} is the worst and Molmo \citep{deitke2025molmo} is consistently the best. Based on \Cref{fig:vlm-hp-all}, we observe that in general, promptable representations aggregated by averaging next-to-last layer prompt embeddings perform the best. From a practical standpoint, this is beneficial, as it eliminates the additional time for answer generation. Ironically, the best prompt is \emph{Do not describe background features. Focus on the robot arm and the [task-obj]}, which explicitly asks VLM to filter out distractors.


\begin{figure*}[t]
    \begin{center}
        \centerline{\includegraphics[width=\textwidth]{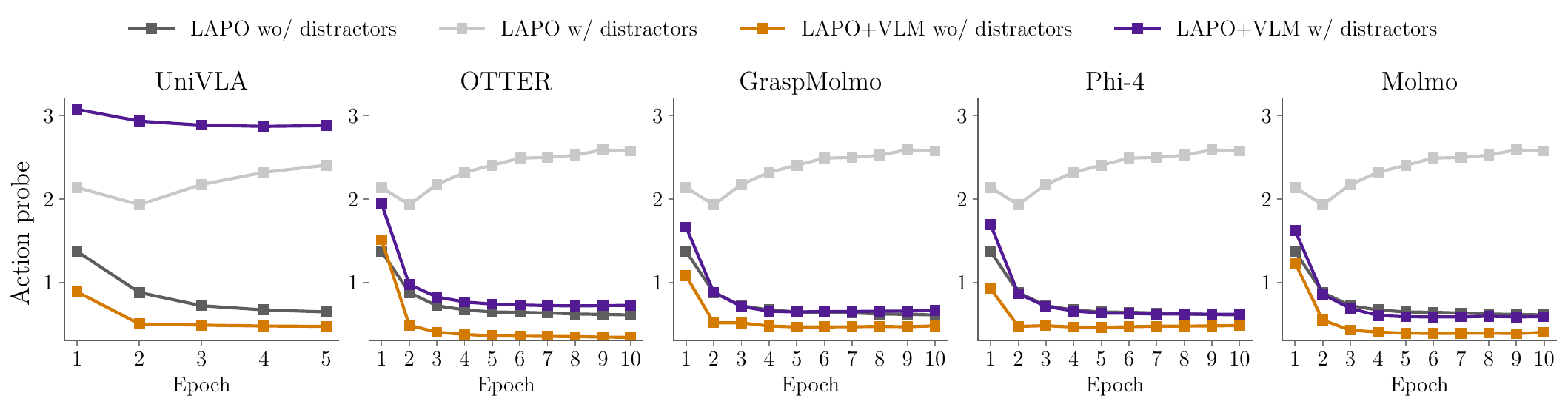}}
        \caption{Action probes comparison for LAPO, baselines and LAPO with promptable representations on full datasets for all tasks in MT10. Results are averaged over three random seeds. As can be seen, promptable representations significantly improves upon LAPO in terms of the latent actions quality, and without any supervision with true actions closes the gap with LAPO without distractors. While all VLMs bring improvements, Molmo achieve best results overall, especially given it high robustness to hyperparameter choices (see \Cref{fig:vlm-benchmark-all}). UniVLA uses two-stage pipeline, each half of total epochs and we visualize the second stage, where task-centric latent actions should be learned. For resulting success rates see \Cref{fig:lapo-vlm-sc}. We provide per-environment probes in \Cref{fig:lapo-vlm-probes-envs}, \Cref{app:add-figs}.
        }
        \label{fig:lapo-vlm-probes}
    \end{center}
    \vskip -0.2in
\end{figure*}



This brings us to a striking conclusion that state-of-the-art VLMs do not necessarily provide better promptable representations. For example, InstructBLIP \citep{dai2023instructblip} outperforms both Gemma-3 \citep{team2025gemma} and Pixtral \citep{agrawal2024pixtral}, despite being considerably older. Furthermore, Cosmos-Reason \citep{azzolini2025cosmos} results indicate that explicit fine-tuning on robotics data is not sufficient to guarantee improved representations. We believe that our results, besides relevance to LAMs, reveal a large blind spot in how VLMs are currently evaluated, with critical implications for robotics and VLA models. 

In addition, DINOv2 and CLIP (see \Cref{fig:vlm-benchmark}) results highlight the vital importance of language conditioning. Due to large-scale pretraining both may possess helpful inductive biases, for example by attending to moving objects, but without language conditioning there is no guarantee that these objects are the ones that are controllable. Both methods achieve the worst latent action quality among all approaches we considered. For instance, OTTER \citep{huang2025otter} uses the same CLIP model but applies simple training-free filtering using text CLIP embeddings. This small modification significantly improves latent action quality, although still not to the level of VLMs such as Molmo.

\textbf{Benchmarking embedding VLMs.} In our main benchmark (see \Cref{fig:vlm-benchmark}), we evaluated conventional VLMs, which were not explicitly trained to produce strong unified representations and therefore required heuristics such as embedding pooling. Recently, a new class of embedding VLMs has emerged \citep{jiang2024vlm2vec,meng2025vlm2vec}. These models are designed specifically to learn high-quality, promptable, and multimodal embeddings for zero-shot retrieval. Given the similarity of their objective to ours, one might expect them to perform better. To test this, we evaluated three recent state-of-the-art models \citep{jiang2024e5,meng2025vlm2vec,gu2025breaking} using the same protocol as earlier, but separately as they require different prompt formats. As can be seen in \Cref{fig:emb-vlm}, such models do not deliver any substantial gains. In fact, VLM2Vec-V2 \citep{meng2025vlm2vec}, best model in its class, performed worse than Gemma-3, which was the weakest model in the main benchmark, and none of the models surpassed Molmo. Our results indicate that embedding VLMs do not actually encode only prompt-specific visual information into the embeddings and fail to deliver the anticipated benefits for our use case. 

\textbf{Why does Molmo perform so well?} Given Molmo’s strong performance, it is natural to ask what drives its improved representations. Directly answering this is difficult, but we can gather indirect evidence suggesting that the gains stem primarily from pre-training data rather than from the specific architecture. Fortunately, Molmo provides two variants: Molmo-D, which uses Qwen2 as its backbone \citep{yang2025qwen3}, and Molmo-O, which uses OLMo \citep{groeneveld2024olmo}, while both employ CLIP \citep{radford2021learning} as the vision encoder. In contrast, Qwen2-VL \citep{wang2024qwen2} does not use CLIP, offering a useful comparison point to disentangle architectural effects. In \Cref{fig:molmo-inv} we show that CLIP alone performs the worst, Molmo-O ranks second after Molmo-D, and Qwen2-VL performs worse still. Since the Molmo variants share the same pre-training data but differ in backbone architecture, we conclude that the likely source of their superior performance lies in the data rather than the architecture. A further hypothesis is that Molmo’s advantage may come from its visual pointing abilities, but this seems unlikely since Moondream-2 also has this ability yet performs worse.


\section{Promptable Representations Unlock Task-Centric Latent Actions}
\label{sec:sc-rate}


\textbf{Setup.} Based on the benchmark results, we selected multiple VLMs for further experiments: Phi-4, Molmo and GraspMolmo. Although all of them achieved improvements in latent action quality upon LAPO on small datasets, it remains necessary to validate whether this performance transfers to the full 5k datasets and yields improved success rates, as this is not guaranteed \citep{nikulin2025latent}. We chose the best hyperparameters for each VLM and trained LAPO+VLM on the full datasets, using three random seeds. As specified in \Cref{sec:setup} we used 16 labeled trajectories with ground-truth actions for final fine-tuning. See \Cref{app:hp} for hyperparameters. 

\textbf{Baselines.} As our primary baselines, we additionally evaluated OTTER \citep{huang2025otter} and UniVLA \citep{bu2025univla}. We note, however, that we are not competing with the VLAs proposed in these works. For UniVLA, we re-implemented their proposed two-stage task-centric LAM training pipeline using our transformer backbone and task instructions provided by MetaWorld \citep{yu2020meta}, which we encoded by a pretrained T5 text encoder \citep{JMLR:v21:20-074}. For OTTER, we adopted their CLIP training-free text-aware visual feature extraction approach to construct a target for FDM in LAPO. All stages other than latent action learning are unified (e.g. share BC actor architecture) across all considered methods to ensure a fair comparison.

\begin{figure}[t]
\vskip 0.2in
    \centering
    \includegraphics[width=\linewidth]{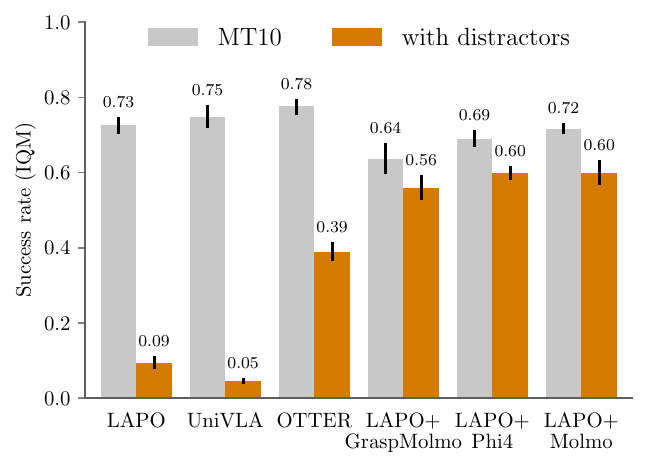}
    \caption{Success rate on MT10 for LAPO and LAPO+VLMs, which uses promptable representations. We use three random seeds and report IQM and $95\%$-CI based on stratified bootstrapping, following the \citet{agarwal2021deep}.}
    \label{fig:lapo-vlm-sc}
\vskip -0.2in
\end{figure}

\textbf{Results.} We present the resulting action probes in \Cref{fig:lapo-vlm-probes} and the final success rates after fine-tuning on real actions in \Cref{fig:lapo-vlm-sc}. As shown in \Cref{fig:lapo-vlm-probes}, LAPO with promptable representations achieves a substantial improvement in latent action quality, both with and without distractors. In the presence of distractors, it nearly closes the gap to LAPO trained without distractors, and without distractors, it slightly outperforms it (e.g., Molmo).

Among the baselines, only UniVLA does not show any improvement in the presence of distractors. As we discussed in \Cref{sec:related}, UniVLA adds a language task instruction embedding to the IDM and FDM inputs, which may help disentangle task-relevant videos from noise at a high level but does not provide the per-step learning signal required to accurately recover low-level ground-truth actions. As a result, during single-task learning the task instruction remains a constant vector, and UniVLA effectively reduces to LAPO.


Crucially, the improvement in action probes on full datasets carries over to downstream performance (see \Cref{fig:lapo-vlm-sc}): success rates increase by a factor of six at max in the presence of distractors, while remaining almost unchanged without them. Interestingly, we found Phi-4 to outperform GraspMolmo, despite having worse probes on small datasets. On full datasets (\Cref{fig:lapo-vlm-probes}), however, Phi-4 is better. This indicates, that while results from a small dataset may carry over to a larger one with some error, probes on the full dataset predict the final success rate with high accuracy. Overall, our results confirm the viability of promptable representations as a clean target for latent action modeling in the presence of distractors.

\begin{figure}[t]
\vskip 0.2in
    \centering
    \includegraphics[width=\linewidth]{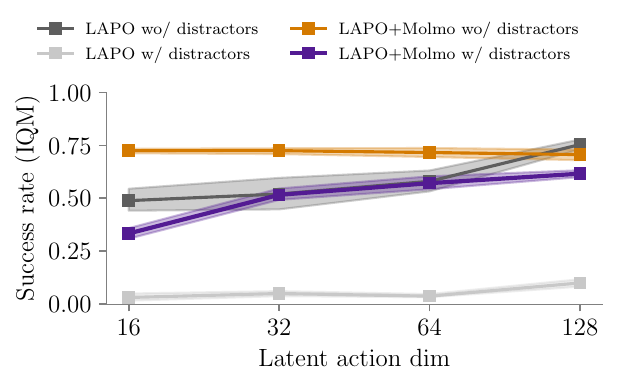}
    \caption{Success rate on MT10 for LAPO and LAPO+Molmo, with varying latent action dimension to control the information bottleneck. We use three random seeds and report IQM and $95\%$-CI based on stratified bootstrapping, following the \citet{agarwal2021deep}.}
    \label{fig:vlm-ld}
\vskip -0.2in
\end{figure}

\textbf{Varying the latent action information bottleneck.} In all previous experiments, we fixed the latent action dimension to 128 to ensure a consistent level of latent action minimality across all methods. However, this choice was inherited from LAPO and may not be optimal. Moreover, as shown by LAPA \citep{ye2024latent}, smaller latent action spaces make the perception-and-language-to-action generation problem significantly easier to learn. We therefore conducted additional experiments that vary the latent action dimension, using the full datasets and three random seeds. The results are summarized in \Cref{fig:vlm-ld}. We observe that promptable representations not only increase success rates in the presence of distractors, but also significantly outperform LAPO without distractors under stronger minimality constraints, such as 16 action dimensions. These results provide further evidence that VLMs help filter out information irrelevant to controllable changes, enabling a more compact latent action space.

\section{Conclusion}

In this work, we demonstrated that promptable representations provided by Vision-Language Models can effectively filter out action-correlated distractors, enabling task-centric latent actions. Our experiments on the Distracting MetaWorld benchmark confirmed that using task-centric promptable representations as targets for LAPO substantially improves both latent action quality and downstream success rates. We hope that our results will inspire the community to explore promptable representations at scale for the next generation of Vision-Language-Action models. We discuss the limitations of our work in the \Cref{app:lim}.

\section*{Impact Statement}

This paper presents work whose goal is to advance the field of Machine
Learning. There are many potential societal consequences of our work, none
which we feel must be specifically highlighted here.





\bibliography{icml2026}
\bibliographystyle{icml2026}

\newpage
\appendix
\onecolumn




\section{Discussion and Limitations}
\label{app:lim}

\textbf{Segmentation, while simple, is not enough.} The concept of extracting VLM embeddings with the hope that they will filter out distractors may initially seem strange. If the goal is to filter out distractors, would not it be more straightforward to simply segment the relevant parts and train LAPO directly in image space using masks? In fact, our benchmark includes VLMs capable of segmentation, such as Sa2VA \citep{yuan2025sa2va}, and we even utilize such prompts (see \Cref{app:tab:prompts}), yet we still rely on embeddings instead of masks. While segmentation is appealing, it does not address the fundamental problem. Consider a scenario with a robotic arm and varying lighting conditions. Even if we segment the arm, we will still get changes in our observations that are not related to the actual actions, such as color shifts and reflections on the arm. The same issue arises with camera movement and changes in perspective. The key, therefore, is to work in a semantic latent space, which is where the common-sense reasoning capabilities of VLMs become crucial.

\textbf{On the choice of MetaWorld benchmark.} One notable limitation of our study is its small scale, as we rely on MetaWorld as our primary benchmark and do not extend our analysis to large VLAs and datasets, such as Open-X \citep{o2024open}. However, this choice is deliberate for two reasons. First, while MetaWorld is simple, with distractors, it is difficult enough to completely break traditional LAMs and to distinguish different VLMs in terms of the promptable representations' quality (as we demonstrate in \Cref{sec:vlm-bench}). As an early exploration, it was crucial to expand in variety (e.g., exploring more VLMs) within our limited resources. We hope that our analysis provides practitioners with valuable insights into the available options. Second, encoding entire datasets is both expensive and time-consuming, as it involves inference with large VLMs (e.g., 8B parameters) and generating answers. For our 5k trajectory datasets, the process can quickly exceed 24 hours, let alone for truly large datasets. Since this is purely inference and gradients are not required, the process can be significantly accelerated, for example, using vLLM \citep{kwon2023efficient}. However, we have left this as future work.

\clearpage
\section{Additional Figures}
\label{app:add-figs}

\begin{figure}[h]
    \begin{center}
        \centerline{\includegraphics[width=0.75\textwidth]{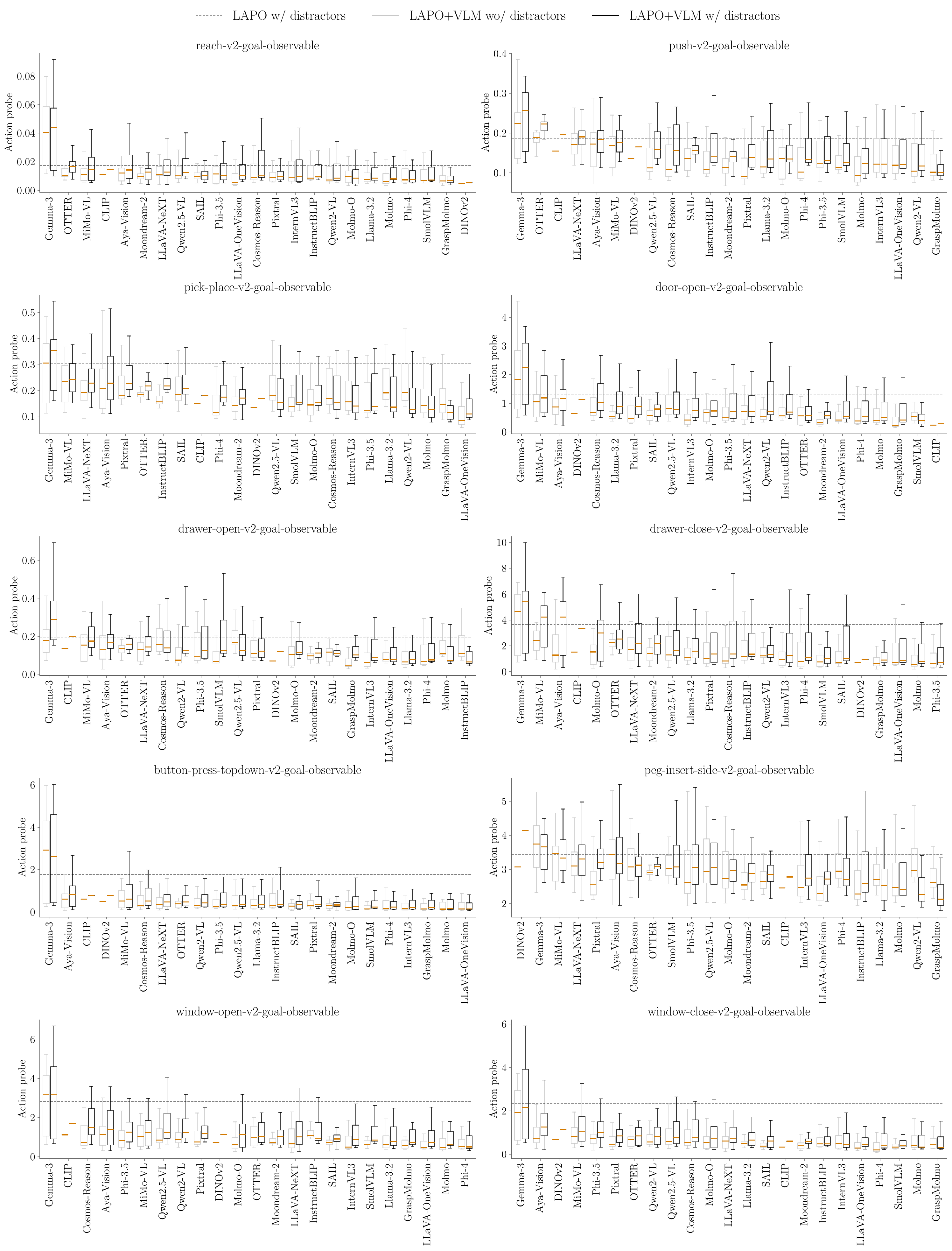}}
        \caption{Aggregation over all hyperparameters for each task in MT10.}
        \label{fig:}
    \end{center}
    \vskip -0.2in
\end{figure}

\begin{figure}[h]
    \begin{center}
        \centerline{\includegraphics[width=0.75\textwidth]{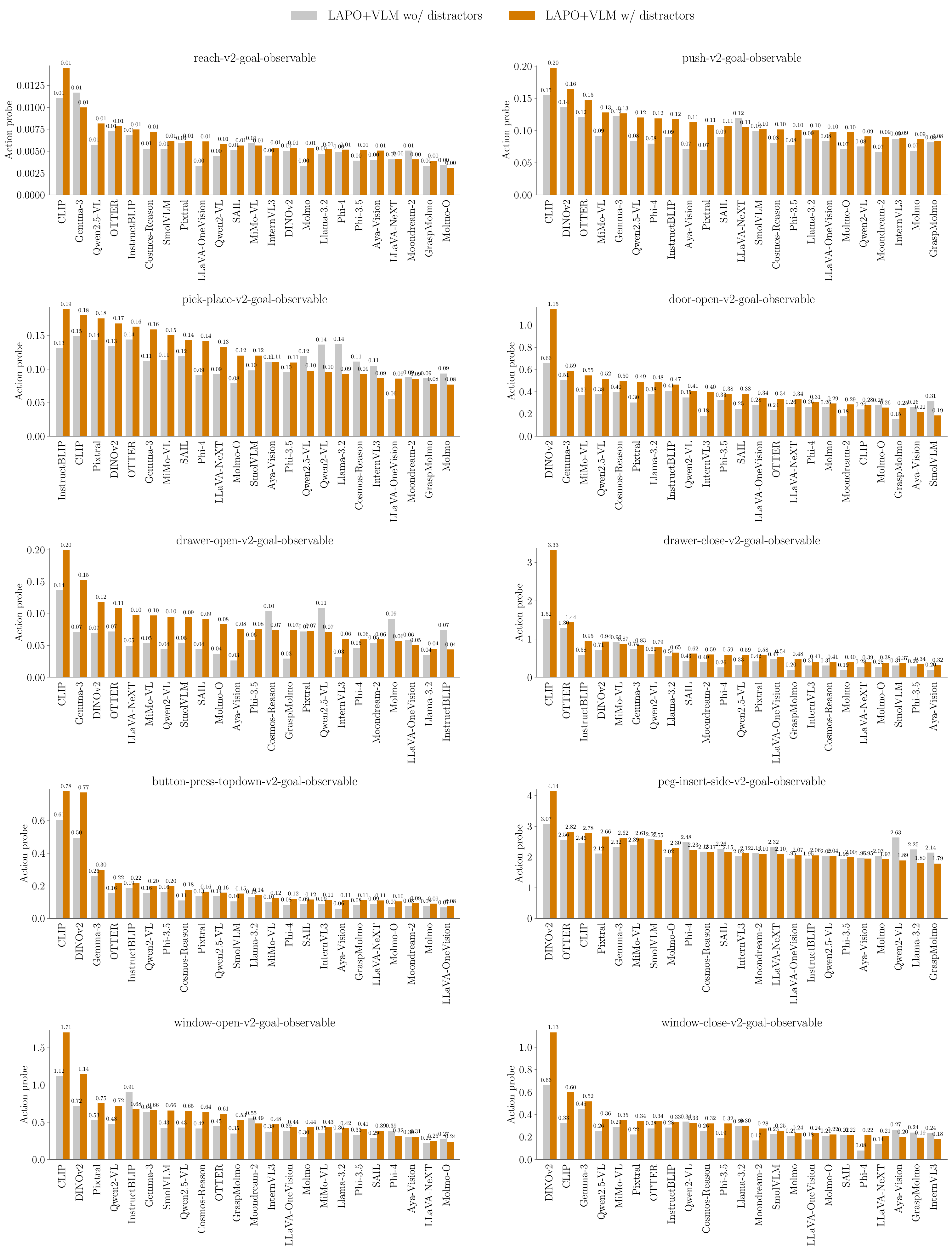}}
        \caption{Probe values for best hyperparameters for each task in MT10.}
        \label{fig:}
    \end{center}
    \vskip -0.2in
\end{figure}

\begin{figure}[t]
    \centering
    \begin{subfigure}[b]{0.8\textwidth}
        \centering
        \centerline{\includegraphics[width=\columnwidth]{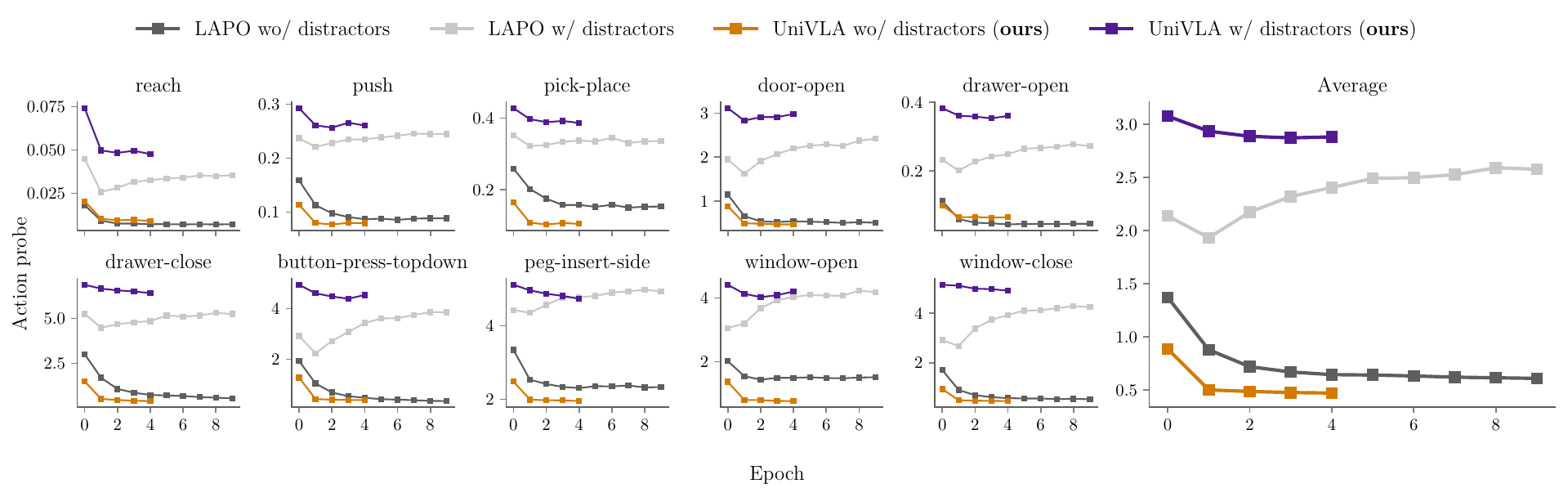}}
        \label{fig:lapo-vlm-probes-envs-univla}
    \end{subfigure}
    
    \begin{subfigure}[b]{0.8\textwidth}
        \centering
        \centerline{\includegraphics[width=\columnwidth]{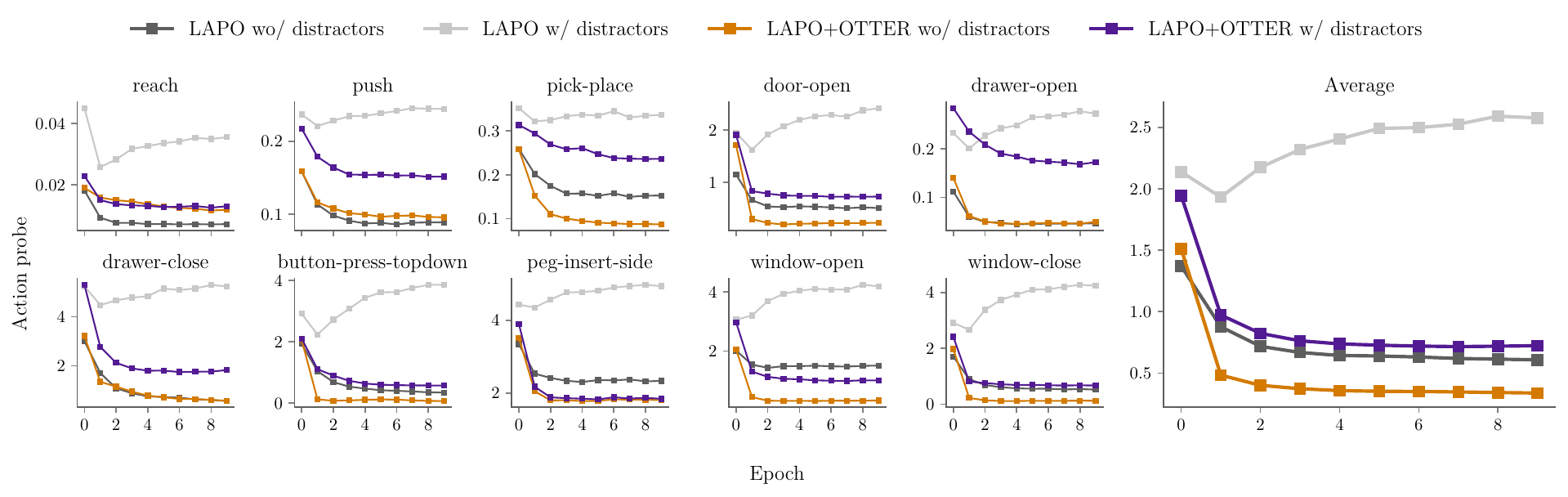}}
        \label{fig:lapo-vlm-probes-envs-otter}
    \end{subfigure}
    \begin{subfigure}[b]{0.8\textwidth}
        \centering
        \centerline{\includegraphics[width=\columnwidth]{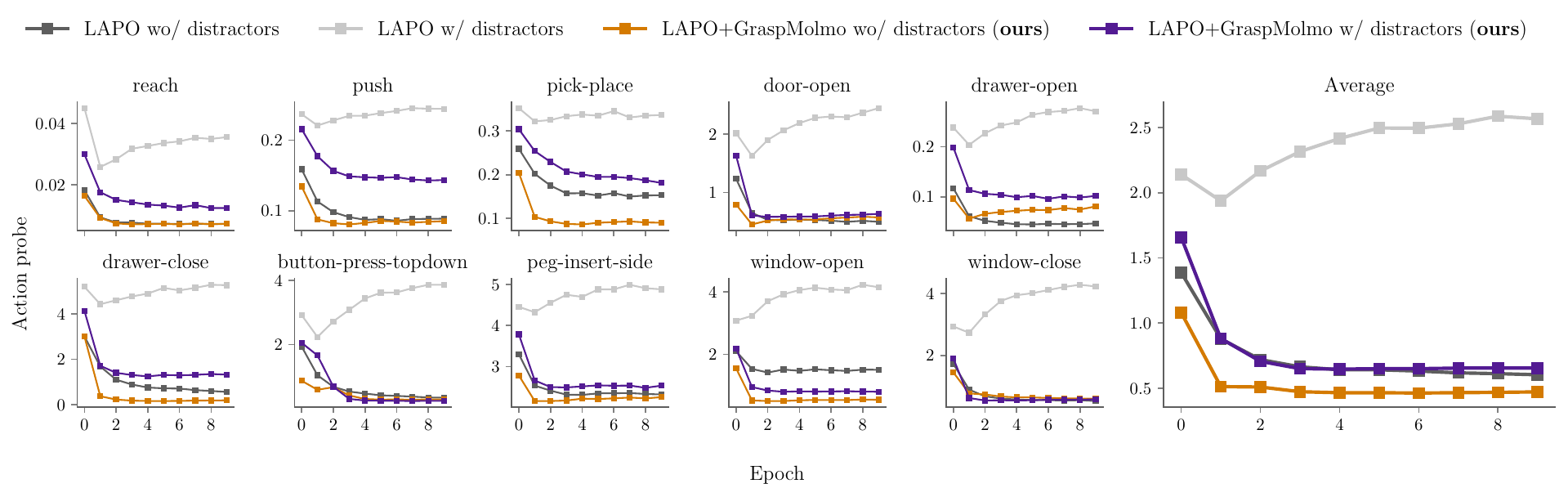}}
        \label{fig:lapo-vlm-probes-envs-grasp-molmo}
    \end{subfigure}
    \begin{subfigure}[b]{0.8\textwidth}
        \centering
        \centerline{\includegraphics[width=\columnwidth]{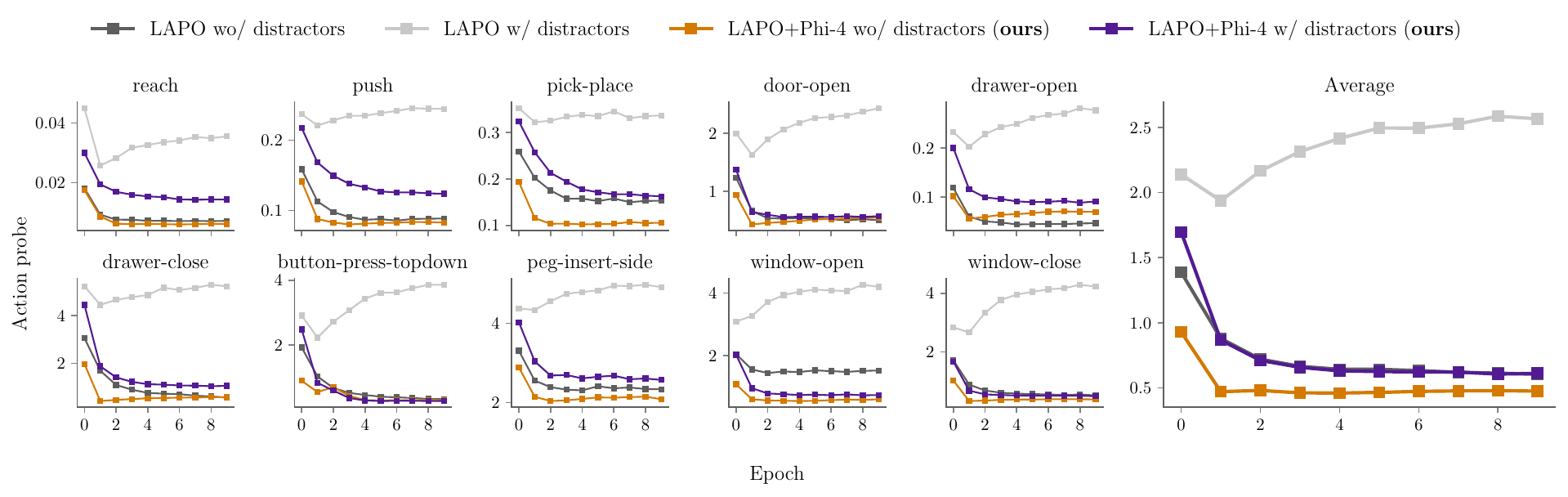}}
        \label{fig:lapo-vlm-probes-envs-phi}
    \end{subfigure}
    \begin{subfigure}[b]{0.8\textwidth}
        \centering
        \centerline{\includegraphics[width=\columnwidth]{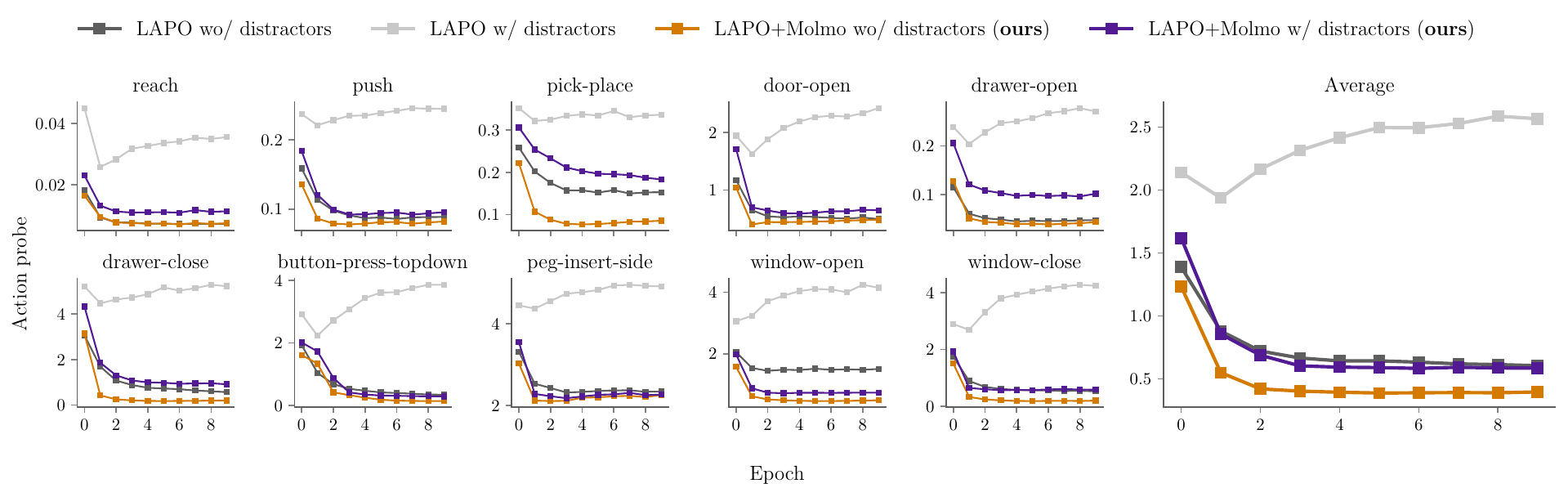}}
        \label{fig:lapo-vlm-probes-envs-molmo}
    \end{subfigure}
    \caption{Action probes comparison for LAPO and LAPO+VLMs on full datasets for all tasks in MT10. Results are averaged over three random seeds. As can be seen, LAPO+VLM significantly improves upon LAPO in terms of the latent actions quality, and without any supervision \textcolor{blue} {with true actions} closes the gap with LAPO without distractors. While all VLMs bring improvements, Molmo achieve best results overall. For resulting success rates see \Cref{fig:lapo-vlm-sc}}
    \label{fig:lapo-vlm-probes-envs}
\end{figure}



\newpage

\begin{figure}[t]
    \begin{center}
        \centerline{\includegraphics[width=\columnwidth]{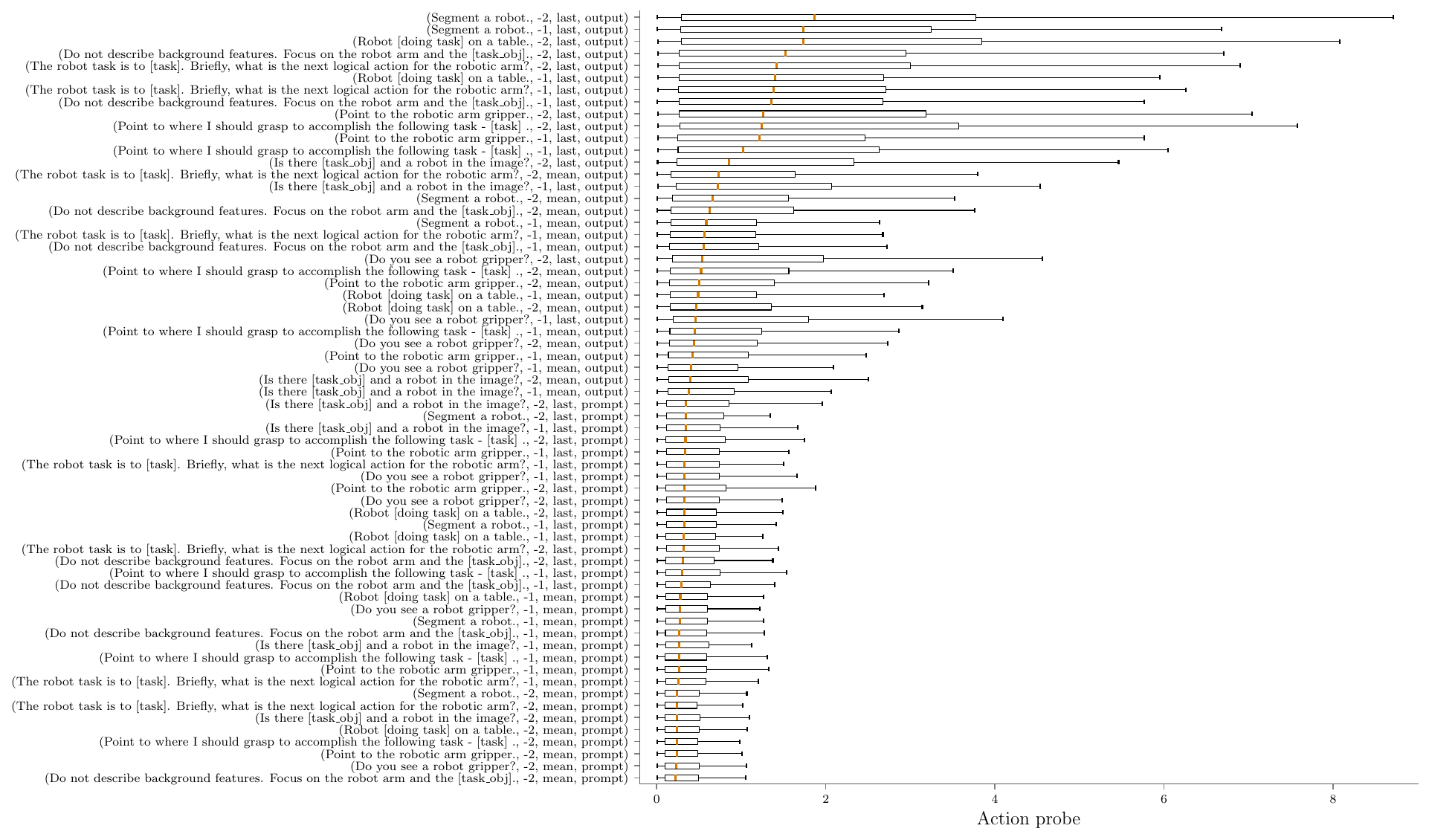}}
        \caption{Action probe rankings across all explored hyperparameter combinations. Smaller probe is better. Values are averaged over all VLMs, tasks and settings, e.g. with and without distractors. Hyperparameters in order: prompt, layer, reduction type and source of the embeddings. Feel free to zoom in!}
        \label{fig:vlm-hp-all}
    \end{center}
\end{figure}

\begin{figure}[h]
    \vskip -0.2in
    \begin{center}
        \centerline{\includegraphics[width=\textwidth]{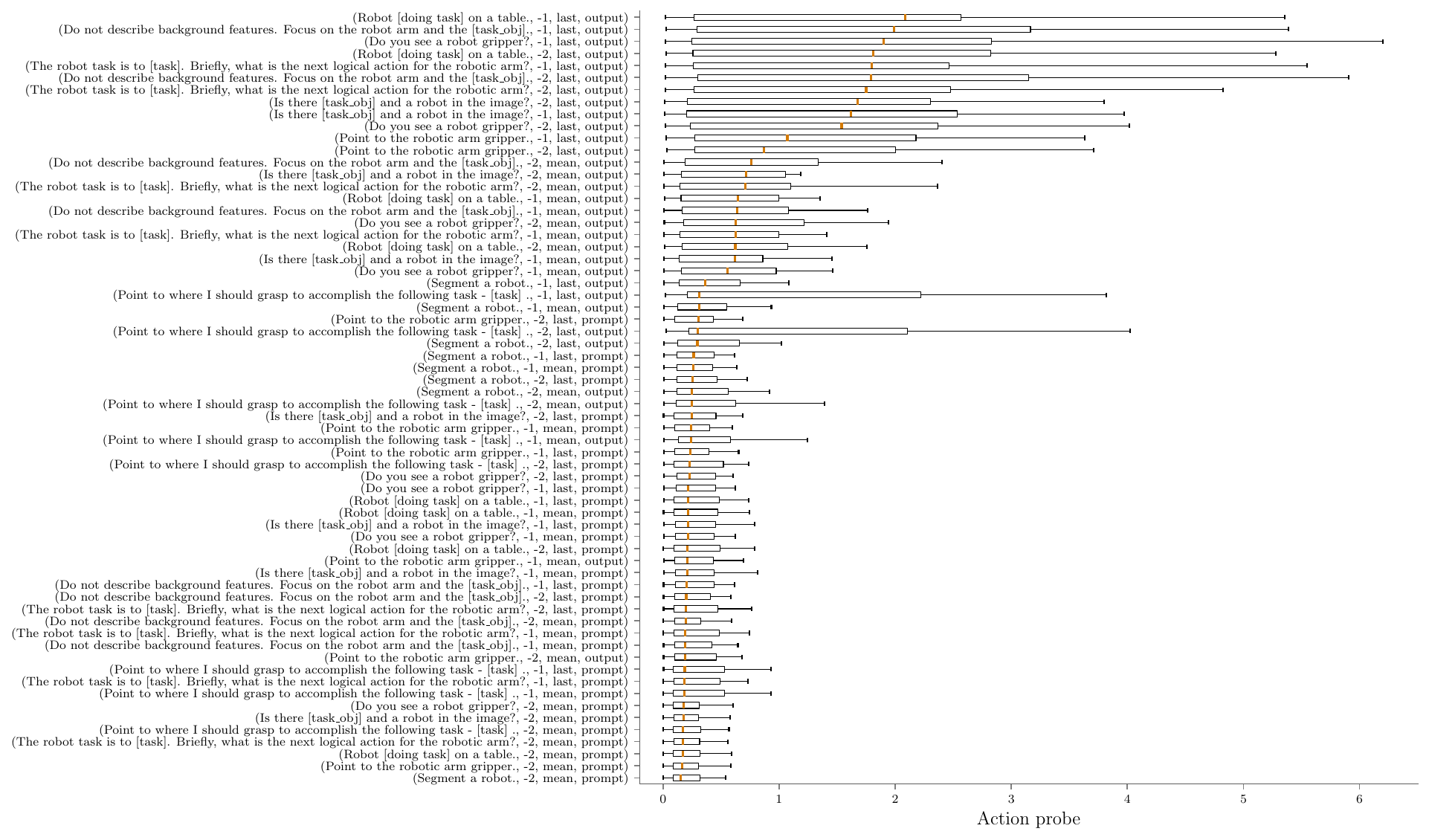}}
        \caption{Action probes ranking for all combinations of hyperparameters explored for Molmo VLM. Values are averaged over all tasks and settings, e.g. with and without distractors.}
        \label{fig:vlm-hp-molmo}
    \end{center}
    \vskip -0.2in
\end{figure}

\clearpage

\section{Vision-Language Models Details}
\label{app:vlm-list}

\begin{table}[h]
\caption{Prompt templates used in our experiments. We adapt them to each task by inserting information relevant to the task. All VLMs explored share the same prompts per task.}
\label{app:tab:prompts}
\begin{center}
\begin{tabular}{l}
\toprule
\textbf{Prompt} \\
\midrule
The robot task is to [task]. Briefly, what is the next logical action for the robotic arm? \\
Do not describe background features. Focus on the robot arm and the [task-obj]. \\
Do you see a robot gripper? \\
Is there [task-obj] and a robot in the image? \\
Robot [doing task] on a table. \\
Point to the robotic arm gripper. \\
Point to where I should grasp to accomplish the following task - [task]. \\
Segment a robot. \\
\bottomrule
\end{tabular}
\end{center}
\end{table}

\begin{table}[h]
\caption{Exact HuggingFace IDs for all VLMs we used. We shortened their names in Figures to save some space.}
\begin{center}
\begin{tabular}{l|l}
\toprule
\textbf{Name} & \textbf{HuggingFace ID} \\
\midrule
InstructBLIP & Salesforce/instructblip-vicuna-7b \\
Molmo  & allenai/Molmo-7B-D-0924 \\
Gemma-3 & google/gemma-3-12b-it \\
Llama-3.2 & unsloth/Llama-3.2-11B-Vision-Instruct \\
Qwen2.5-VL & Qwen/Qwen2.5-VL-7B-Instruct \\
InternVL3 & OpenGVLab/InternVL3-8B \\
Cosmos-Reason & nvidia/Cosmos-Reason1-7B \\ 
Phi-4 & microsoft/Phi-4-multimodal-instruct \\
LLaVA-OneVision & llava-hf/llava-onevision-qwen2-7b-ov-hf \\
SmolVLM & HuggingFaceTB/SmolVLM2-2.2B-Instruct \\
Pixtral & mistral-community/pixtral-12b \\
GraspMolmo & allenai/GraspMolmo \\
Qwen2-VL & Qwen/Qwen2-VL-7B-Instruct \\
Phi-3.5 & Lexius/Phi-3.5-vision-instruct \\
LLaVA-NeXT & llava-hf/llava-v1.6-mistral-7b-hf \\
Aya-Vision & CohereLabs/aya-vision-8b \\
MiMo-VL & XiaomiMiMo/MiMo-VL-7B-RL \\
Molmo-O & allenai/Molmo-7B-O-0924 \\
SAIL & ByteDance-Seed/SAIL-7B \\
Moondream-2 & vikhyatk/moondream2 \\
VLM2Vec & VLM2Vec/VLM2Vec-V2.0 \\
UniME & DeepGlint-AI/UniME-LLaVA-1.6-7B \\
E5-V & royokong/e5-v \\
Janus-Pro & deepseek-community/Janus-Pro-7B \\
Sa2VA & ByteDance/Sa2VA-4B \\
\bottomrule
\end{tabular}
\end{center}
\end{table}

\clearpage
\section{Hyperparameters}
\label{app:hp}

\begin{table}[h]
    \caption{LAPO-ResNet hyperparameters. Names are exactly follow the configuration files used in code.}
    \label{sample-table}
    \vskip 0.15in
    \begin{center}
    \begin{tabular}{l|l|r}
    \toprule
    \textbf{Part} & \textbf{Parameter} & \textbf{Value} \\
    \midrule
    \multirow{6}{*}{General} 
        & frame\_stack & 4 \\
        & probe\_learning\_rate & 0.0003 \\
        & disable\_distractors & True \\
        & seed & 0 \\
        & eval\_seed & 0 \\
        & eval\_episodes & 50 \\
    \midrule
    \multirow{13}{*}{Latent action learning}
        & latent\_action\_dim & 128 \\
        & idm\_encoder\_scale & 5 \\
        & idm\_encoder\_num\_res\_blocks & 1 \\
        & idm\_encoder\_channels & [16, 16, 32, 32, 128, 128, 256] \\
        & fdm\_encoder\_scale & 1 \\
        & fdm\_encoder\_num\_res\_blocks & 1 \\
        & fdm\_encoder\_channels & [16, 16, 32, 32, 128, 128, 256] \\
        & num\_epochs & 10 \\
        & batch\_size & 64 \\
        & learning\_rate & 0.0001 \\
        & weight\_decay & 0.0 \\
        & warmup\_epochs & 1 \\
        & grad\_norm & - \\
    \midrule
    \multirow{8}{*}{Latent behavior cloning} 
        & num\_epochs & 10 \\
        & batch\_size & 64 \\
        & learning\_rate & 0.0001 \\
        & weight\_decay & 0.0 \\
        & warmup\_epochs & 0 \\
        & encoder\_scale & 5 \\
        & encoder\_num\_res\_blocks & 1 \\
        & encoder\_channels & [16, 16, 32, 32, 128, 128, 256] \\
    \midrule
    \multirow{5}{*}{Latent actions decoding} 
        & total\_updates & 100000 \\
        & batch\_size & 64 \\
        & learning\_rate & 0.001 \\
        & hidden\_dim & 128 \\
        & num\_labeled\_trajectories & [16, 8, 2, 4] \\
    \bottomrule
    \end{tabular}
    \end{center}
\end{table}   

\begin{table}[h]
    \caption{LAPO-Trans hyperparameters. Names exactly follow the configuration files used in code.}
    \label{tab:hyperparams}
    \vskip 0.15in
    \begin{center}
    \begin{tabular}{l|l|r}
    \toprule
    \textbf{Part} & \textbf{Parameter} & \textbf{Value} \\
    \midrule
    \multirow{6}{*}{General} 
        & frame\_stack & 4 \\
        & probe\_learning\_rate & 0.0003 \\
        & disable\_distractors & True \\
        & seed & 0 \\
        & eval\_seed & 0 \\
        & eval\_episodes & 50 \\
    \midrule
    \multirow{17}{*}{Latent action learning}
        & latent\_action\_dim & 128 \\
        & patch\_size & 32 \\
        & fdm\_use\_cross\_attn & False \\
        & idm\_hidden\_dim & 896 \\
        & idm\_num\_layers & 4 \\
        & idm\_num\_heads & 16 \\
        & fdm\_hidden\_dim & 256 \\
        & fdm\_num\_layers & 4 \\
        & fdm\_num\_heads & 8 \\
        & normalize\_qk & False \\
        & pre\_norm & True \\
        & num\_epochs & 10 \\
        & batch\_size & 64 \\
        & learning\_rate & 0.0001 \\
        & weight\_decay & 0.0 \\
        & warmup\_epochs & 1 \\
        & grad\_norm & - \\
    \midrule
    \multirow{8}{*}{Latent behavior cloning} 
        & num\_epochs & 10 \\
        & batch\_size & 64 \\
        & learning\_rate & 0.0001 \\
        & weight\_decay & 0.0 \\
        & warmup\_epochs & 0 \\
        & encoder\_scale & 5 \\
        & encoder\_num\_res\_blocks & 1 \\
        & encoder\_channels & [16, 16, 32, 32, 128, 128, 256] \\
    \midrule
    \multirow{5}{*}{Latent actions decoding} 
        & total\_updates & 100000 \\
        & batch\_size & 64 \\
        & learning\_rate & 0.001 \\
        & hidden\_dim & 128 \\
        & num\_labeled\_trajectories & [16, 8, 2, 4] \\
    \bottomrule
    \end{tabular}
    \end{center}
\end{table}

\begin{table}[h]
    \caption{LAPO+VLM hyperparameters. Names exactly follow the configuration files used in code.}
    \label{tab:hyperparams-vlm}
    \vskip 0.15in
    \begin{center}
    \begin{tabular}{l|l|r}
    \toprule
    \textbf{Part} & \textbf{Parameter} & \textbf{Value} \\
    \midrule
    \multirow{8}{*}{General} 
        & frame\_stack & 4 \\
        & probe\_learning\_rate & 0.0003 \\
        & disable\_distractors & True \\
        & seed & 0 \\
        & eval\_seed & 0 \\
        & eval\_episodes & 50 \\
    \midrule
    \multirow{5}{*}{VLM (example)}
        & type & molmo \\
        & prompt & Point to the robotic arm gripper. \\
        & layer & 27 \\
        & target & output \\
        & reduce\_strategy & mean \\
    \midrule
    \multirow{10}{*}{Latent action learning}
        & latent\_action\_dim & 128 \\
        & idm\_encoder\_scale & 5 \\
        & idm\_encoder\_num\_res\_blocks & 1 \\
        & idm\_encoder\_channels & [16, 16, 32, 32, 128, 128, 256] \\
        & fdm\_hidden\_dim & 1024 \\
        & fdm\_num\_layers & 4 \\
        & fdm\_expand & 4 \\
        & num\_epochs & 200 \\
        & batch\_size & 64 \\
        & learning\_rate & 0.0001 \\
        & weight\_decay & 0.0 \\
        & warmup\_epochs & 1 \\
        & grad\_norm & - \\
    \midrule
    \multirow{7}{*}{Latent behavior cloning} 
        & num\_epochs & 10 \\
        & batch\_size & 64 \\
        & learning\_rate & 0.0001 \\
        & weight\_decay & 0.0 \\
        & warmup\_epochs & 0 \\
        & encoder\_scale & 5 \\
        & encoder\_num\_res\_blocks & 1 \\
        & encoder\_channels & [16, 16, 32, 32, 128, 128, 256] \\
    \midrule
    \multirow{5}{*}{Latent actions decoding} 
        & total\_updates & 100000 \\
        & batch\_size & 64 \\
        & learning\_rate & 0.001 \\
        & hidden\_dim & 128 \\
        & num\_labeled\_trajectories & [16, 8, 2, 4] \\
    \bottomrule
    \end{tabular}
    \end{center}
\end{table}

\end{document}